\pdfoutput=1

\documentclass[11pt]{article}

\usepackage{acl}

\usepackage{times}
\usepackage{latexsym}

\usepackage[T1]{fontenc}

\usepackage[utf8]{inputenc}

\usepackage{microtype}

\usepackage{multirow}

\usepackage{subcaption}
\usepackage{graphicx}
\usepackage{multirow}
\usepackage{paralist}
\usepackage{hyperref}
\usepackage{arydshln}
\definecolor{ForestGreen}{RGB}{34,139,34}

\title{BanglaBERT: Language Model Pretraining and Benchmarks for Low-Resource Language Understanding Evaluation in Bangla}

\author{
Abhik Bhattacharjee$^1$\thanks{~ These authors contributed equally to this work.} , Tahmid Hasan$^1$\footnotemark[1] , Wasi Uddin Ahmad$^2$\thanks{~ Work done while at UCLA.},  Kazi Samin$^1$,\\
\textbf{Md Saiful Islam}$^3$, \textbf{Anindya Iqbal}$^1$, \textbf{M. Sohel Rahman}$^1$, \textbf{Rifat Shahriyar}$^1$\\ [3pt]
Bangladesh University of Engineering and Technology (BUET)$^1$,\\AWS AI Labs$^2$, University of Rochester$^3$\\
[3pt]
\texttt{abhik@ra.cse.buet.ac.bd}, \texttt{\{tahmidhasan,rifat\}@cse.buet.ac.bd}\\
}

\date{}

\begin{document}
\maketitle

\begin{abstract}

In this work, we introduce \emph{BanglaBERT}, a BERT-based Natural Language Understanding (NLU) model pretrained in Bangla, a widely spoken yet low-resource language in the NLP literature. To pretrain BanglaBERT, we collect 27.5 GB of Bangla pretraining data (dubbed `Bangla2B+') by crawling 110 popular Bangla sites. We introduce two downstream task datasets on natural language inference and question answering and benchmark on four diverse NLU tasks covering text classification, sequence labeling, and span prediction. In the process, we bring them under the first-ever Bangla Language Understanding Benchmark (BLUB). BanglaBERT achieves state-of-the-art results outperforming multilingual and monolingual models. We are making the models, datasets, and a leaderboard publicly available at \url{https://github.com/csebuetnlp/banglabert} to advance Bangla NLP.

\end{abstract}
\section{Introduction}

Despite being the sixth most spoken language in the world with over 300 million native speakers constituting 4\% of the world's total population,\footnote{\url{https://w.wiki/Psq}} Bangla is considered a resource-scarce language. \citet{joshi-etal-2020-state} categorized Bangla in the language group that lacks efforts in labeled data collection and relies on self-supervised pretraining \citep{devlin-etal-2019-bert, radford2019language, liu2019roberta} to boost the natural language understanding (NLU) task performances. To date, the Bangla language has been continuing to rely on fine-tuning multilingual pretrained language models (PLMs) \citep{ashrafi2020banner, das-etal-2021-emotion, islam-etal-2021-sentnob-dataset}. However, since multilingual PLMs cover a wide range of languages \citep{conneau2019cross, conneau2019unsupervised}, they are large (have hundreds of millions of parameters) and require substantial computational resources for fine-tuning. They also tend to show degraded performance for low-resource languages \citep{wu-dredze-2020-languages} on downstream NLU tasks. Motivated by the triumph of language-specific models (\citet{martin2019camembert, polignano2019alberto, canete2020spanish, baly2020arabert}, \textit{inter alia}) over multilingual models in many other languages, in this work, we present \textbf{BanglaBERT} -- a BERT-based \citep{devlin-etal-2019-bert} Bangla NLU model pretrained on 27.5 GB data (which we name `\textbf{Bangla2B+}') we meticulously crawled 110 popular Bangla websites to facilitate NLU applications in Bangla. 
Since most of the downstream task datasets for NLP applications are in the English language, to facilitate zero-shot transfer learning between English and Bangla, we additionally pretrain a model in both languages; we name the model \textbf{BanglishBERT}.



We also introduce two datasets on Bangla Natural Language Inference (NLI) and Question Answering (QA), tasks previously unexplored in Bangla, and evaluate both pretrained models on four diverse downstream tasks on sentiment classification, NLI, named entity recognition, and QA. We bring these tasks together to establish the first-ever \textbf{B}angla \textbf{L}anguage \textbf{U}nderstanding \textbf{B}enchmark (\textbf{BLUB}). We compare widely used multilingual models to BanglaBERT using BLUB and find that both models excel on all the tasks. 

We summarize our contributions as follows:

\begin{compactenum}
    \item We present two new pretrained models: BanglaBERT and BanglishBERT, and introduce new Bangla NLI and QA datasets.
    \item We introduce the Bangla Language Understanding Benchmark (BLUB) and show that, in the supervised setting, BanglaBERT outperforms mBERT and XLM-R (base) by 6.8 and 4.3 BLUB scores, while in zero-shot cross-lingual transfer, BanglishBERT outperforms them by 15.8 and 10.8, respectively.
    \item We provide the code, models, and a leaderboard to spur future research on Bangla NLU.
\end{compactenum}

\section{BanglaBERT}\label{sec:pretraining}


\subsection{Pretraining Data}

A high volume of good quality text data is a prerequisite for pretraining large language models. For instance, BERT \citep{devlin-etal-2019-bert} is pretrained on the English Wikipedia and the Books corpus \citep{zhu2015aligning} containing 3.3 billion tokens. Subsequent works like RoBERTa \citep{liu2019roberta} and XLNet \citep{yang2019xlnet} used more extensive web-crawled data with heavy filtering and cleaning. 

Bangla is a rather resource-constrained language in the web domain; for example, the Bangla Wikipedia dump from July 2021 is only 650 MB, two orders of magnitudes smaller than the English Wikipedia. As a result, we had to crawl the web extensively to collect our pretraining data. We selected 110 Bangla websites by their Amazon Alexa rankings\footnote{\url{www.alexa.com/topsites/countries/BD}} and the volume and quality of extractable texts by inspecting each website. The contents included encyclopedias, news, blogs, e-books, stories, social media/forums, etc.\footnote{The complete list can be found in the Appendix.} The amount of data totaled around 35 GB. 

There are noisy sources of Bangla data dumps, a couple of prominent ones being OSCAR \citep{suarez2019asynchronous} and CCNet \citep{wenzek2019ccnet}. They contained many offensive texts; we found them infeasible to clean thoroughly. Fearing their potentially harmful impacts \citep{luccioni-viviano-2021-whats}, we opted not to use them. We further discuss ethical considerations at the end of the paper.

\subsection{Pre-processing}

We performed thorough deduplication on the pretraining data, removed non-textual contents (e.g., HTML/JavaScript tags), and filtered out non-Bangla pages using a language classifier \citep{joulin2016bag}. After the processing, the dataset was reduced to 27.5 GB in size containing 5.25M documents having 306.66 words on average.

We trained a Wordpiece \citep{wu2016google} vocabulary of 32k subword tokens on the resulting corpus with a 400 character alphabet, kept larger than the native Bangla alphabet to capture code-switching \citep{poplack1980sometimes} and allow romanized Bangla contents for better generalization. We limited the length of a training sample to 512 tokens and did not cross document boundaries \citep{liu2019roberta} while creating a data point. After tokenization, we had 7.18M samples with an average length of 304.14 tokens and containing 2.18B tokens in total; hence we named the dataset `\textit{Bangla2B+}'.

\subsection{Pretraining Objective}

Self-supervised pretraining objectives leverage unlabeled data.
For example, BERT \citep{devlin-etal-2019-bert} was pretrained with masked language modeling (MLM) and next sentence prediction (NSP). Several works built on top of this, e.g., RoBERTa \citep{liu2019roberta} removed NSP and pretrained with longer sequences, SpanBERT \citep{joshi-etal-2020-spanbert} masked contiguous spans of tokens, while works like XLNet \citep{yang2019xlnet} introduced objectives like factorized language modeling.

We pretrained BanglaBERT using ELECTRA \citep{clark2020electra}, pretrained with the Replaced Token Detection (RTD) objective, where a generator and a discriminator model are trained jointly. The generator is fed as input a sequence with a portion of the tokens masked (15\% in our case) and is asked to predict them using the rest of the input (i.e., standard MLM). The masked tokens are then replaced by tokens sampled from the generator's output distribution for the corresponding masks, and the discriminator then has to predict whether each token is from the original sequence or not. After pretraining, the discriminator is used for fine-tuning. \citet{clark2020electra} argued that RTD back-propagates loss from all tokens of a sequence, in contrast to 15\% tokens of the MLM objective, giving the model more signals to learn from. Evidently, ELECTRA achieves comparable downstream performance to RoBERTa or XLNet with only a quarter of their training time. This computational efficiency motivated us to use ELECTRA for our implementation of BanglaBERT.

\begin{table*}[!tbh]
\centering\setlength{\tabcolsep}{3.5pt}
\begin{tabular}{llrrrll}
\hline
\textbf{Task} & \textbf{Corpus} & \textbf{|Train|} & \textbf{|Dev|} & \textbf{|Test|} & \textbf{Metric} & \textbf{Domain}\\
\hline
Sentiment Classification & SentNoB & 12,575 & 1,567 & 1,567 & Macro-F1 & Social Media\\ 
Natural Language Inference & BNLI & 381,449 & 2,419 & 4,895 & Accuracy & Miscellaneous\\
Named Entity Recognition & MultiCoNER & 14,500 & 800 & 800 & Micro-F1 & Miscellaneous\\
Question Answering & BQA, TyDiQA & 127,771 & 2,502 & 2,504 & EM/F1 & Wikipedia\\
\hline
\end{tabular}
\caption{
Statistics of the Bangla Language Understanding Evaluation (BLUB) benchmark.
}
\label{tab:blub}
\end{table*}

\subsection{Model Architecture \& Hyperparameters} 

We pretrained the base ELECTRA model (a 12-layer Transformer encoder with 768 embedding size, 768 hidden size, 12 attention heads, 3072 feed-forward size, generator-to-discriminator ratio $\frac{1}{3}$, 110M parameters) with 256 batch size for 2.5M steps on a v3-8 TPU instance on GCP. We used the Adam \citep{kingma2014adam} optimizer with a 2e-4 learning rate and linear warmup of 10k steps.

\subsection{BanglishBERT} 

Often labeled data in a low-resource language for a task may not be available but be abundant in high-resource languages like English. In these scenarios, zero-shot cross-lingual transfer \citep{artetxe2019massively} provides an effective way to be still able to train a multilingual model on that task using the high-resource languages and be able to transfer to low-resource ones. To this end, we pretrained a bilingual model, named \textbf{BanglishBERT}, on Bangla and English together using the same set of hyperparameters mentioned earlier. We used the BERT pretraining corpus as the English data and trained a joint bilingual vocabulary (each language having $\sim$16k tokens). We upsampled the Bangla data during training to equal the participation of both languages. 

\section{The Bangla Language Understanding Benchmark (BLUB)}

Many works have studied different Bangla NLU tasks in isolation, e.g., sentiment classification \citep{das2010phrase, sharfuddin2018deep, tripto-bysa-2018}, semantic textual similarity \citep{shajalal2018semantic}, parts-of-speech (PoS) tagging \citep{alam2016bidirectional}, named entity recognition (NER) \citep{ashrafi2020banner}. However, Bangla NLU has not yet had a comprehensive, unified study. Motivated by the surge of NLU research brought about by benchmarks in other languages, e.g.,  English \citep{wang2018glue}, French \citep{le2020flaubert}, Korean \citep{park2021klue}, we establish the first-ever Bangla Language Understanding Benchmark (BLUB). NLU generally comprises three types of tasks: text classification, sequence labeling, and text span prediction. Text classification tasks can further be sub-divided into single-sequence and sequence-pair classification. Therefore, we consider a total of four tasks for BLUB. For each task type, we carefully select one downstream task dataset. We emphasize the quality and open availability of the datasets while making the selection. We briefly mention them below.

\begin{table*}[!tbh]
\centering\setlength{\tabcolsep}{11pt}
\begin{tabular}{l c c c c c c}
\hline
\textbf{Models} & \textbf{|Params.|} & \textbf{SC} & \textbf{NLI} & \textbf{NER} & \textbf{QA} & \textbf{BLUB Score}\\
\hline
\multicolumn{7}{l}{\emph{Zero-shot cross-lingual transfer}} \\
\cdashline{1-7}
mBERT & 180M & 27.05 & 62.22 & 39.27 & 59.01/64.18 & 50.35\\
XLM-R (base) & 270M & 42.03 & 72.18 & 45.37 & 55.03/61.83 & 55.29\\
XLM-R (large) & 550M & 49.49 & 78.13 & 56.48 & 71.13/77.70 & 66.59\\
\textbf{BanglishBERT} & 110M & 48.39 & 75.26 & 55.56 & \textbf{72.87}/\textbf{78.63} & 66.14\\
\cdashline{1-7}
\multicolumn{7}{l}{\emph{Supervised fine-tuning}} \\
\cdashline{1-7}
mBERT & 180M & 67.59 & 75.13 & 68.97 & 67.12/72.64 & 70.29\\
XLM-R (base) & 270M & 69.54 & 78.46 & 73.32 & 68.09/74.27 & 72.82\\
XLM-R (large) & 550M & 70.97 & \textbf{82.40} & \textbf{78.39} & \textbf{73.15}/\textbf{79.06} & 76.79\\
IndicBERT & 18M & 68.41 & 77.11 & 54.13 & 50.84/57.47 & 61.59 \\
sahajBERT & 18M & 71.12 & 76.92 & 70.94 & 65.48/70.69 & 71.03\\
\textbf{BanglishBERT} & 110M & 70.61 & 80.95 & 76.28 & \textbf{72.43}/78.40 & 75.73\\
\textbf{BanglaBERT} & 110M & \textbf{72.89}  & \textbf{82.80} & \textbf{77.78} & \textbf{72.63}/\textbf{79.34} & \textbf{77.09} \\
\hline
\end{tabular}
\caption{Performance comparison of pretrained models on different downstream tasks. Scores in bold texts have statistically significant ($p < 0.05$) difference from others with bootstrap sampling \citep{koehn-2004-statistical}.}
\label{tab:benchmark}
\end{table*}

\paragraph{1. Single-Sequence Classification}
Sentiment classification is perhaps the most-studied Bangla NLU task, with some of the earlier works dating back over a decade \citep{das2010phrase}. Hence, we chose this as the single-sequence classification task. However, most Bangla sentiment classification datasets are not publicly available. We could only find two public datasets: \textit{BYSA} \citep{tripto-bysa-2018} and \textit{SentNoB} \citep{islam-etal-2021-sentnob-dataset}. We found BYSA to have many duplications. Even worse, many duplicates had different labels. SentNoB had better quality and covered a broader set of domains, making the classification task more challenging. Hence, we opted to use the latter.

\paragraph{2. Sequence-pair Classification}
In contrast to single-sequence classification, there has been a dearth of sequence-pair classification works in Bangla. We found work on semantic textual similarity \citep{shajalal2018semantic}, but the dataset is not publicly available. As such, we curated a new Bangla Natural Language Inference (\textit{BNLI}) dataset for sequence-pair classification. We chose NLI as the representative task due to its fundamental importance in NLU. Given two sentences, a premise and a hypothesis as input, a model is tasked to predict whether or not the hypothesis is entailment, contradiction, or neutral to the premise. We used the same curation procedure as the XNLI \citep{conneau2018xnli} dataset: we translated the MultiNLI \citep{williams2017broad} training data using the English to Bangla translation model by \citet{hasan2020not} and had the evaluation sets translated by expert human translators.\footnote{More details are presented in the ethical considerations section.} Due to the possibility of the incursion of errors during automatic translation, we used the Language-Agnostic BERT Sentence Embeddings (LaBSE) \citep{feng2020language} of the translations and original sentences to compute their similarity and discarded all sentences below a similarity threshold of 0.70. Moreover, to ensure good-quality human translation, we used similar quality assurance strategies as \citet{guzman2019flores}.

\paragraph{3. Sequence Labeling} 
In this task, all words of a text sequence have to be classified. Named Entity Recognition (NER) and Parts-of-Speech (PoS) tagging are two of the most prominent sequence labeling tasks. We chose the Bangla portion of SemEval 2022 \textit{MultiCoNER} \citep{multiconer-data} dataset for BLUB.

\paragraph{4. Span Prediction}
Extractive question answering is a standard choice for text span prediction. Similar to BNLI, we machine-translated the \textit{SQuAD 2.0} \citep{rajpurkar-etal-2018-know} dataset and used it as the training set (BQA). For validation and test, We used the Bangla portion of the \textit{TyDiQA}\footnote{We removed the Yes/No questions from TyDiQA and subsampled the unanswerable questions to have equal proportion.} \citep{clark-etal-2020-tydi} dataset. We posed the task analogous to SQuAD 2.0: presented with a text passage and a question, a model has to predict whether or not it is answerable. If answerable, the model has to find the minimal text span that answers the question.

We present detailed statistics of the BLUB benchmark in Table \ref{tab:blub}.

\section{Experiments \& Results}

\paragraph{Setup}

We fine-tuned BanglaBERT and BanglishBERT on the four downstream tasks and compared them with several multilingual models: mBERT \citep{devlin-etal-2019-bert}, XLM-R base and large \citep{conneau2019unsupervised}, and IndicBERT \citep{kakwani2020indicnlpsuite}, a multilingual model for Indian languages; and sahajBERT \citep{diskin2021distributed}, an ALBERT-based \citep{Lan2020ALBERT:} PLM for Bangla. All pretrained models were fine-tuned for 3-20 epochs with batch size 32, and the learning rate was tuned from \{2e-5, 3e-5, 4e-5, 5e-5\}. The final models were selected based on the validation performances after each epoch. We performed fine-tuning with three random seeds and reported their average scores in Table \ref{tab:benchmark}. We reported the average performance of all tasks as the BLUB score. 

\paragraph{Zero-shot Transfer}
We show the zero-shot cross-lingual transfer results of the multilingual models fine-tuned on the English counterpart of each dataset (SentNob has no English equivalent; hence we used the Stanford Sentiment Treebank \citep{socher2013recursive} for the sentiment classification task) in Table \ref{tab:benchmark}. In zero-shot transfer setting, BanglishBERT achieves strong cross-lingual performance over similar-sized models and falls marginally short of XLM-R (large). This is an expected outcome since cross-lingual effectiveness depends explicitly on model size \citep{K2020Cross-Lingual}.

\paragraph{Supervised Fine-tuning}
In the supervised fine-tuning setup, BanglaBERT outperformed multilingual models and monolingual sahajBERT on all the tasks, achieving a BLUB score of 77.09, even coming head-to-head with XLM-R (large). On the other hand, BanglishBERT marginally lags behind BanglaBERT and XLM-R (large).
BanglaBERT is not only superior in performance but also substantially compute- and memory-efficient. For instance, it may seem that sahajBERT is more efficient than BanglaBERT due to its smaller size, but it takes 2-3.5x time and 2.4-3.33x memory as BanglaBERT to fine-tune.
\footnote{We present a detailed comparison in the Appendix.}


\begin{figure}[!tbh]
\centering
\includegraphics[width=0.47\textwidth]{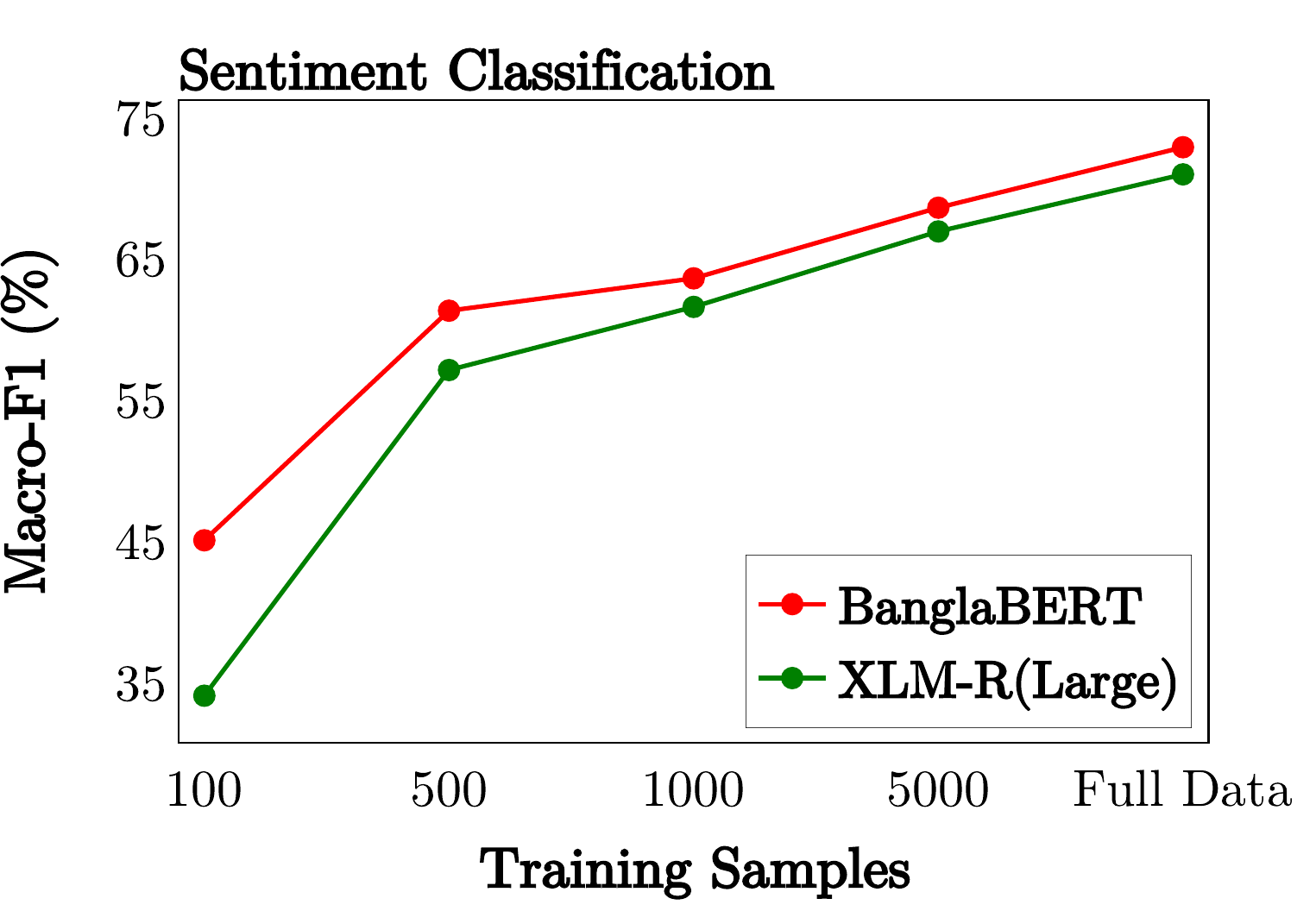}
\vspace{0.0mm}
\includegraphics[width=0.47\textwidth]{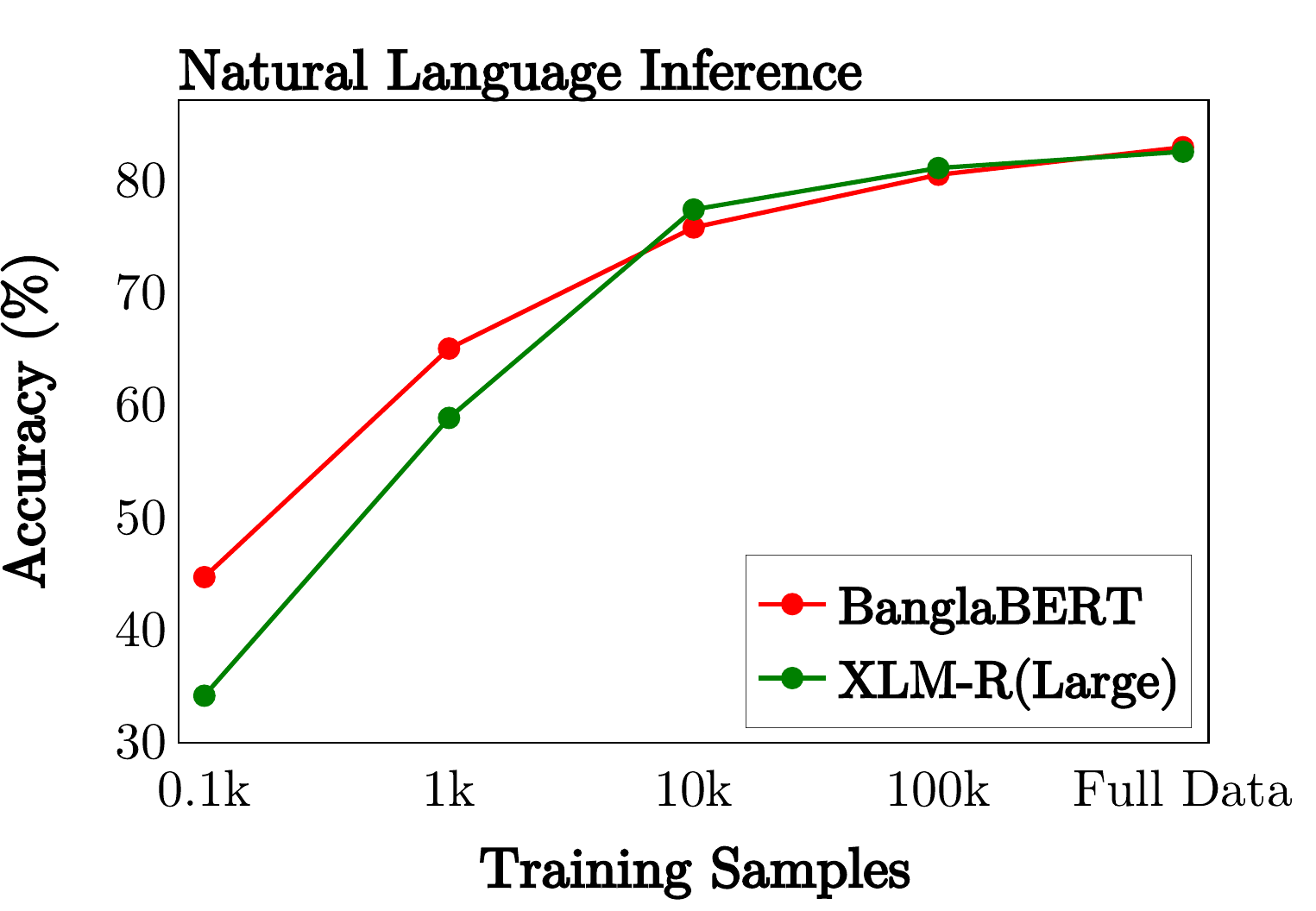}
\caption{Sample-efficiency tests with SC and NLI.}
\label{fig:sample_complexity_nli}
\end{figure}

\paragraph{Sample efficiency}
It is often challenging to annotate training samples in real-world scenarios, especially for low-resource languages like Bangla.
So, in addition to compute- and memory-efficiency, sample-efficiency \citep{howard2018universal} is another necessity of PLMs. To assess the sample efficiency of BanglaBERT, we limit the number of training samples and see how it fares against other models. We compare it with XLM-R (large) and plot their performances on the SC and NLI tasks\footnote{Results for the other tasks can be found in the Appendix.} for different sample size in Figure \ref{fig:sample_complexity_nli}. 

Results show that when we have fewer number of samples ($\leq 1k$), BanglaBERT has substantially better performance (2-9\% on SC and 6-10\% on NLI with $p < 0.05$) over XLM-R (large), making it more practically applicable for resource-scarce downstream tasks.
\section{Conclusion \& Future Works}
Creating language-specific models is often infeasible for low-resource languages lacking ample data. Hence, researchers are compelled to use multilingual models for languages that do not have strong pretrained models.
To this end, we introduced BanglaBERT and BanglishBERT, two NLU models in Bangla, a widely spoken yet low-resource language. We presented new downstream datasets on NLI and QA, and established the BLUB benchmark, setting new state-of-the-art results with BanglaBERT. 
In future, we will include other Bangla NLU benchmarks (e.g., dependency parsing \citep{de-marneffe-universal-2021}) in BLUB and investigate the benefits of initializing Bangla NLG models from BanglaBERT.

\section*{Acknowledgements}
We would like to thank the Research and Innovation Centre for Science and Engineering (RISE), BUET, for funding the project, and Intelligent Machines Limited and Google TensorFlow Research Cloud (TRC) Program for providing cloud support. 

\section*{Ethical Considerations}

\paragraph{Dataset and Model Release} The \emph{Copy Right Act, 2000}\footnote{\url{http://bdlaws.minlaw.gov.bd/act-details-846.html}} of Bangladesh allows reproduction and public release of copy-right materials for non-commercial research purposes. As a transformative research work, we will release BanglaBERT under a non-commercial license. Furthermore, we will release only the pretraining data for which we know the distribution will not cause any copyright infringement. The downstream task datasets can all be made publicly available under a similar non-commercial license.

\paragraph{Quality Control in Human Translation} 
Translations were done by expert translators who provide translation services for renowned Bangla newspapers. Each translated sentence was further assessed for quality by another expert. If found to be of low quality, it was again translated by the original translator. The sample was then discarded altogether if found to be of low quality again. Fewer than 100 samples were discarded in this process. Translators were paid as per standard rates in local currencies. 

\paragraph{Text Content} 
We tried to minimize offensive texts in the pretraining data by explicitly crawling the sites where such contents would be nominal. However, we cannot guarantee that there is absolutely no objectionable content present and therefore recommend using the model carefully, especially for text generation purposes. 

\bibliographystyle{acl_natbib}
\bibliography{anthology,arr2022}

\clearpage

\section*{Appendix}

\subsection*{Pretraining Data Sources}

We used the following sites for data collection. We categorize the sites into six types:\\

\noindent\textbf{Encyclopedia:}

\begin{itemize}\itemsep-2mm
    \item \href{https://bn.banglapedia.org/}{bn.banglapedia.org}
    \item \href{https://bn.wikipedia.org/}{bn.wikipedia.org}
    \item \href{https://songramernotebook.com/}{songramernotebook.com}
\end{itemize}

\noindent\textbf{News:}

\begin{itemize}\itemsep-2mm
    \item \href{https://www.anandabazar.com/}{anandabazar.com}
    \item \href{https://www.arthoniteerkagoj.com/}{arthoniteerkagoj.com}
    \item \href{https://www.bangla.24livenewspaper.com/}{bangla.24livenewspaper.com}
    \item \href{https://bangla.bdnews24.com/}{bangla.bdnews24.com}
    \item \href{https://bangla.dhakatribune.com/}{bangla.dhakatribune.com}
    \item \href{https://bangla.hindustantimes.com/}{bangla.hindustantimes.com}
    \item \href{http://bangladesherkhela.com/}{bangladesherkhela.com}
    \item \href{https://www.banglanews24.com/}{banglanews24.com}
    \item \href{https://www.banglatribune.com/}{banglatribune.com}
    \item \href{https://www.bbc.com/bengali}{bbc.com}
    \item \href{https://www.bd-journal.com/}{bd-journal.com}
    \item \href{https://www.bd-pratidin.com/}{bd-pratidin.com}
    \item \href{https://www.bd24live.com/bangla/}{bd24live.com}
    \item \href{https://bengali.indianexpress.com/}{bengali.indianexpress.com}
    \item \href{http://www.bigganprojukti.com/biggan-projukti/}{bigganprojukti.com}
    \item \href{https://bonikbarta.net/}{bonikbarta.net}
    \item \href{http://chakarianews.com/}{chakarianews.com}
    \item \href{https://www.channelionline.com/}{channelionline.com}
    \item \href{https://ctgtimes.com/}{ctgtimes.com}
    \item \href{https://www.ctn24.com/}{ctn24.com}
    \item \href{https://www.daily-bangladesh.com/}{daily-bangladesh.com}
    \item \href{https://www.dailyagnishikha.com/}{dailyagnishikha.com}
    \item \href{https://dainikazadi.net/}{dainikazadi.net}
    \item \href{https://dainikdinkal.net/}{dainikdinkal.net}
    \item \href{https://dailyfulki.com/}{dailyfulki.com}
    \item \href{https://www.dailyinqilab.com/}{dailyinqilab.com}
    \item \href{https://www.dailynayadiganta.com/}{dailynayadiganta.com}
    \item \href{https://dailysangram.com/}{dailysangram.com}
    \item \href{https://dailysylhet.com/}{dailysylhet.com}
    \item \href{https://www.dainikamadershomoy.com/}{dainikamadershomoy.com}
    \item \href{https://www.dainikshiksha.com/}{dainikshiksha.com}
    \item \href{https://dhakardak-bd.com/}{dhakardak-bd.com}
    \item \href{https://dmpnews.org/}{dmpnews.org}
    \item \href{https://www.dw.com/bn/}{dw.com}
    \item \href{https://eisamay.indiatimes.com/}{eisamay.indiatimes.com}
    \item \href{https://www.ittefaq.com.bd/}{ittefaq.com.bd}
    \item \href{https://www.jagonews24.com/}{jagonews24.com}
    \item \href{https://www.jugantor.com/}{jugantor.com}
    \item \href{https://www.kalerkantho.com/}{kalerkantho.com}
    \item \href{https://www.manobkantha.com.bd/}{manobkantha.com.bd}
    \item \href{https://mzamin.com/}{mzamin.com}
    \item \href{https://www.ntvbd.com/}{ntvbd.com}
    \item \href{https://onnodristy.com/}{onnodristy.com}
    \item \href{https://pavilion.com.bd/}{pavilion.com.bd}
    \item \href{https://www.prothomalo.com/}{prothomalo.com}
    \item \href{https://www.protidinersangbad.com/}{protidinersangbad.com}
    \item \href{https://www.risingbd.com/}{risingbd.com}
    \item \href{https://www.rtvonline.com/}{rtvonline.com}
    \item \href{https://samakal.com/}{samakal.com}
    \item \href{https://www.sangbadpratidin.in/}{sangbadpratidin.in}
    \item \href{https://www.somoyerkonthosor.com/}{somoyerkonthosor.com}
    \item \href{https://www.somoynews.tv/}{somoynews.tv}
    \item \href{https://www.tbsnews.net/bangla/}{tbsnews.net}
    \item \href{http://www.teknafnews.com/}{teknafnews.com}
    \item \href{https://www.thedailystar.net/bangla/}{thedailystar.net}
    \item \href{https://www.voabangla.com/}{voabangla.com}
    \item \href{https://zeenews.india.com/bengali}{zeenews.india.com}
    \item \href{https://zoombangla.com/}{zoombangla.com}
\end{itemize}

\noindent\textbf{Blogs:}

\begin{itemize}\itemsep-2mm
    \item \href{https://www.amrabondhu.com/}{amrabondhu.com}
    \item \href{http://banglablog.in/}{banglablog.in}
    \item \href{https://bigganblog.org/}{bigganblog.org}
    \item \href{https://biggani.org/}{biggani.org}
    \item \href{https://bigyan.org.in/}{bigyan.org.in}
    \item \href{https://www.bishorgo.com/}{bishorgo.com}
    \item \href{https://cadetcollegeblog.com/}{cadetcollegeblog.com}
    \item \href{http://www.choturmatrik.com/}{choturmatrik.com}
    \item \href{https://horoppa.wordpress.com/}{horoppa.wordpress.com}
    \item \href{https://muktangon.blog/}{muktangon.blog}
    \item \href{https://roar.media/bangla}{roar.media/bangla}
    \item \href{http://www.sachalayatan.com/}{sachalayatan.com}
    \item \href{https://shodalap.org/}{shodalap.org}
    \item \href{http://shopnobaz.net/}{shopnobaz.net}
    \item \href{https://www.somewhereinblog.net/}{somewhereinblog.net}
    \item \href{http://subeen.com/}{subeen.com}
    \item \href{https://tunerpage.com/}{tunerpage.com}
    \item \href{https://tutobd.com/}{tutobd.com}
\end{itemize}

\noindent\textbf{E-books/Stories:}

\begin{itemize}\itemsep-2mm
    \item \href{https://banglaepub.github.io/}{banglaepub.github.io}
    \item \href{https://bengali.pratilipi.com/}{bengali.pratilipi.com}
    \item \href{https://bn.wikisource.org/wiki/}{bn.wikisource.org}
    \item \href{https://www.ebanglalibrary.com/}{ebanglalibrary.com}
    \item \href{https://eboipotro.github.io/}{eboipotro.github.io}
    \item \href{https://golpokobita.com/}{golpokobita.com}
    \item \href{https://www.kaliokalam.com/}{kaliokalam.com}
    \item \href{https://shirisherdalpala.net/}{shirisherdalpala.net}
    \item \href{https://www.tagoreweb.in/}{tagoreweb.in}
\end{itemize}

\noindent\textbf{Social Media/Forums:}

\begin{itemize}\itemsep-2mm
    \item \href{http://www.banglacricket.com/}{banglacricket.com}
    \item \href{https://bn.globalvoices.org/}{bn.globalvoices.org}
    \item \href{https://helpfulhub.com/}{helpfulhub.com}
    \item \href{https://www.nirbik.com/}{nirbik.com}
    \item \href{https://www.pchelplinebd.com/}{pchelplinebd.com}
    \item \href{https://www.techtunes.io/}{techtunes.io}
\end{itemize}

\noindent\textbf{Miscellaneous:}

\begin{itemize}\itemsep-2mm
    \item \href{https://banglasonglyric.com/}{banglasonglyric.com}
    \item \href{http://bdlaws.minlaw.gov.bd/}{bdlaws.minlaw.gov.bd}
    \item \href{https://www.bdup24.com/}{bdup24.com}
    \item \href{https://bengalisongslyrics.com/}{bengalisongslyrics.com}
    \item \href{http://dakghar.org/}{dakghar.org}
    \item \href{https://www.gdn8.com/}{gdn8.com}
    \item \href{https://gunijan.org.bd/}{gunijan.org.bd}
    \item \href{https://www.hrw.org/bn/}{hrw.org}
    \item \href{https://jakir.me/}{jakir.me}
    \item \href{http://www.jhankarmahbub.com/}{jhankarmahbub.com}
    \item \href{https://www.jw.org/bn/}{jw.org}
    \item \href{https://www.lyricsbangla.com/}{lyricsbangla.com}
    \item \href{http://www.neonaloy.com}{neonaloy.com}
    \item \href{https://porjotonlipi.com/}{porjotonlipi.com}
    \item \href{https://www.sasthabangla.com/author/sasthabangla/}{sasthabangla.com}
    \item \href{https://tanzil.net/trans/}{tanzil.net}
\end{itemize}

We wrote custom crawlers for each site above (except the Wikipedia dumps). 

\subsection*{Additional Sample Efficiency Tests}

\begin{figure}[!tbh]
 \centering
 \includegraphics[width=0.48\textwidth]{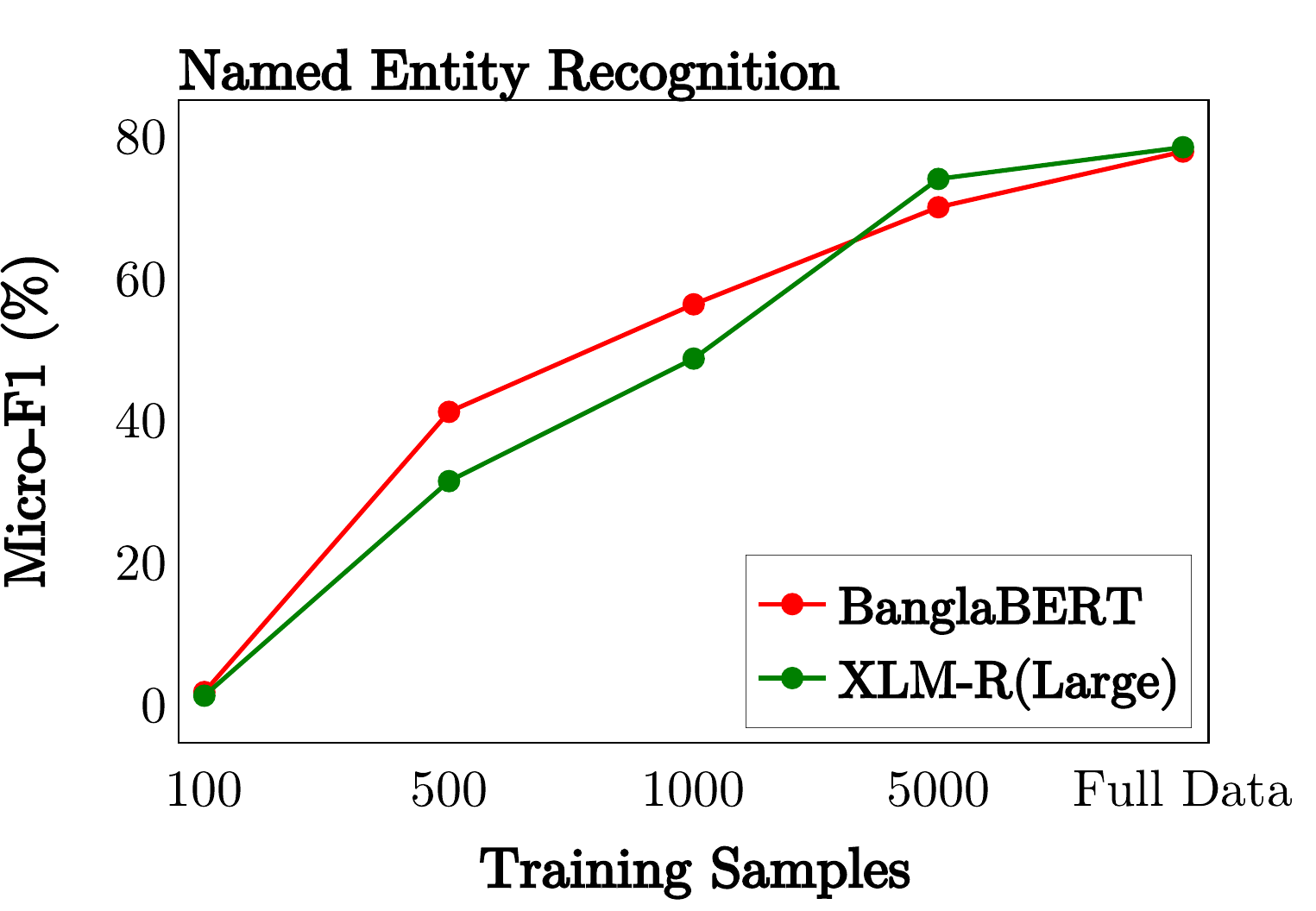}
 \vspace{0.0mm}
 \includegraphics[width=0.48\textwidth]{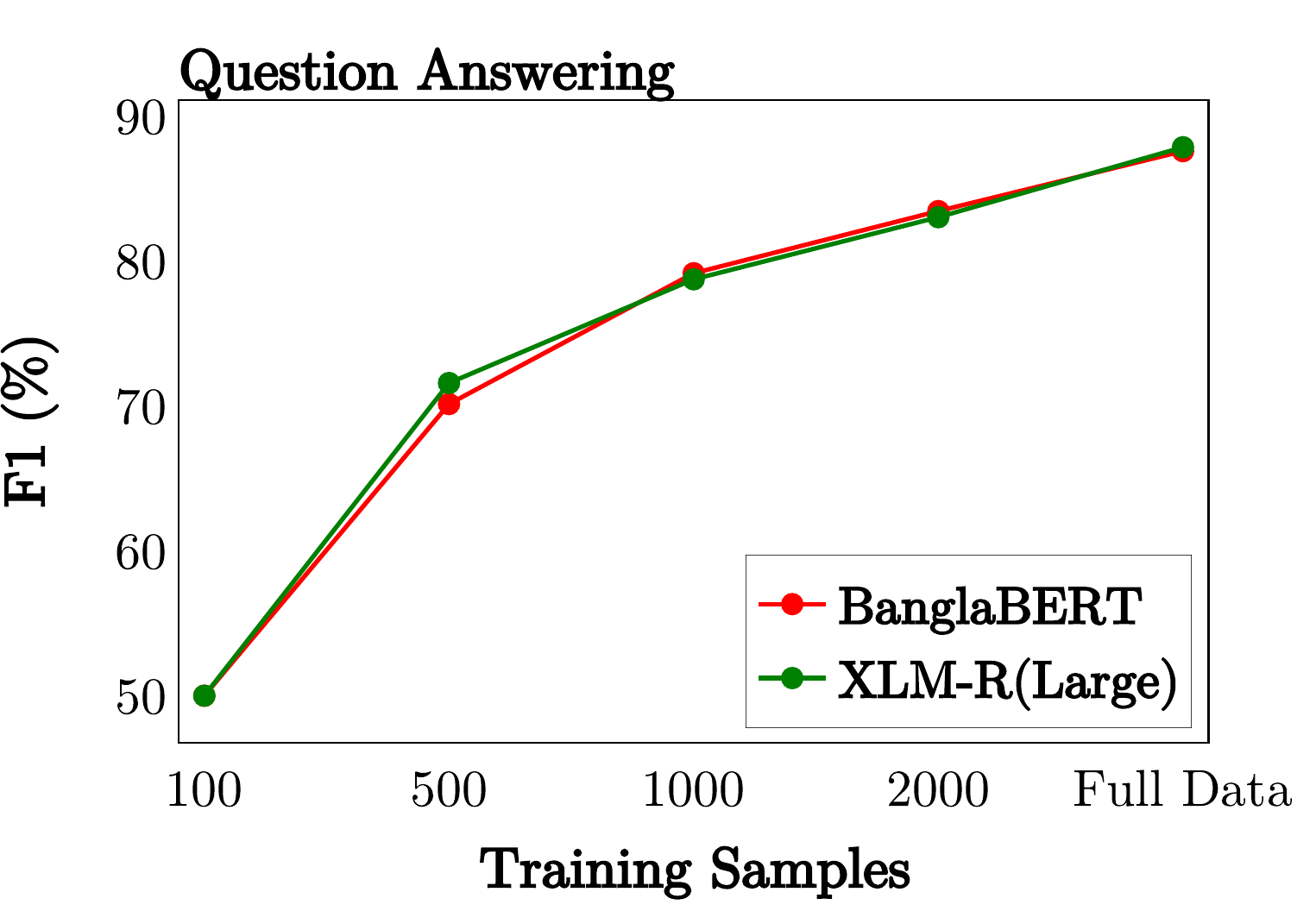}
 \caption{Sample-efficiency tests with NER and QA.}
 \label{fig:sample_complexity_qa}
\end{figure}

We plot the the sample efficiency results of the NER and QA tasks in Figure \ref{fig:sample_complexity_qa}.

Similar results are also observed here for the NER task, where BanglaBERT is more sample-efficient when we have $\leq 1k$ training samples. In the QA task however, both models have identical performance for all sample counts.

\subsection*{Compute and Memory Efficiency Tests}
To validate that BanglaBERT is more efficient in terms of memory and compute, we measured each model's training time and memory usage during the fine-tuning of each task. All tests were done on a desktop machine with an 8-core Intel Core-i7 11700k CPU and NVIDIA RTX 3090 GPU. We used the same batch size, gradient accumulation steps, and sequence length for all models and tasks for a fair comparison. We use relative time and memory (GPU VRAM) usage considering those of BanglaBERT as units. The results are shown in Table \ref{tab:time-mem}. (We mention the upper and lower values of the different tasks for each model)

\begin{table}[!tbh]
\centering
\resizebox{\linewidth}{!}
{%
\begin{tabular}{lcc}
\hline
\textbf{Model} & \textbf{Time} & \textbf{Memory Usage}\\
\hline
mBERT & 1.14x-1.92x & 1.12x-2.04x\\
XLM-R (base) & 1.29-1.81x & 1.04-1.63x\\
XLM-R (large) & 3.81-4.49x & 4.44-5.55x\\
SahajBERT & 2.40-3.33x & 2.07-3.54x\\
BanglaBERT & \textbf{1.00x} & \textbf{1.00x}\\
\hline
\end{tabular}
}
\caption{Compute and memory efficiency tests}\label{tab:time-mem}
\end{table}

\end{document}